\documentclass[10pt,twocolumn,letterpaper]{article}

\usepackage{cvpr}
\usepackage{times}
\usepackage[pdftex]{graphicx}
\usepackage{amsmath}
\usepackage{amssymb}
\usepackage{booktabs}
\usepackage{bm}
\usepackage[pagebackref]{hyperref}
\hypersetup{breaklinks=true}
\usepackage{xurl}
\begin{document}

% 学習データの属性ラベルが不要で公平な顔認証の距離学習手法
\title{LabellessFace: Fair Metric Learning for Face Recognition\\without Attribute Labels}

\author{
Tetsushi Ohki$^{1,2}$,
Yuya Sato$^{1}$,
Masakatsu Nishigaki$^{1}$,
Koichi Ito$^{3}$\\
{$^1$Shizuoka University, Shizuoka, JP}, 
{$^2$RIKEN AIP, Tokyo, JP},
{$^3$Tohoku University, Miyagi, JP}\\
{\tt\small \{ohki@, sato@sec, nisigaki@\}.inf.shizuoka.ac.jp}, 
{\tt\small ito@aoki.ecei.tohoku.ac.jp}
}

\maketitle
\thispagestyle{empty}

\begin{abstract}
% 機械学習システムは，膨大なアプリケーションへの適用が進み，その社会的な影響から，公平性の側面が脚光を浴び始めている．
% As Machine learning systems are increasingly used in wide range of applications, the aspect of fairness is beginning to come into focus due to its social implications.
%
% 1つだけではないので，複数のうちの1つとしました．
% Demographic bias is a major challenge for face recognition systems.
%-------------------------------
% 顔認証においても，認証対象の人種や性別などの属性の間の識別性能が一貫しないことが課題とされてきた．
% 問題：多くの研究では人種(ある一属性)については公平な学習手法を考案しているが，他の属性については公平ではない
% In face recognition, the inconsistency of discrimination performance across different attributes of the subjects, such as race and gender, has been identified as a challenge.
%
% 公平な識別性能を実現するモデルの学習手法に関するこれまでの研究の問題点は，認証対象が持つ様々な属性の中から，特定の属性（主に人種）だけを対象にしており，他の属性については考慮していないことである．
% なぜかというと，それらの研究はデータセットに付与された属性ラベルを利用して学習をしているから
% Existing methods have been limited by a predominant focus on specific demographic group, particulary race, while overlooking others.
%
% それらの研究の多くは，属性のラベル付け作業が必要なことから大規模なデータセットにおける適用を困難とする．さらに識別性能の公平性がデータセットに付与された属性ラベルに深く依存しており，ラベリングされていない属性に対しては対処できない．
% この属性ラベルのアノテーションは，複数の属性をもれなく正確に行うことが難しい
% 個人間で公平なモデルであれば，どのような属性間でも公平であるから，本研究では個人愛で公平なモデルを学習する方針を取ります，そのために個人間でマージンを最適にします．ということを言わないと問題→提案手法の間に飛躍がある

Demographic bias is one of the major challenges for face recognition systems.
The majority of existing studies on demographic biases are heavily dependent on specific demographic groups or demographic classifier, making it difficult to address performance for unrecognised groups.
This paper introduces ``LabellessFace'', a novel framework that improves demographic bias in face recognition without requiring demographic group labeling typically required for fairness considerations. We propose a novel fairness enhancement metric called the class favoritism level, which assesses the extent of favoritism towards specific classes across the dataset. Leveraging this metric, we introduce the fair class margin penalty, an extension of existing margin-based metric learning. This method dynamically adjusts learning parameters based on class favoritism levels, promoting fairness across all attributes. By treating each class as an individual in facial recognition systems, we facilitate learning that minimizes biases in authentication accuracy among individuals.
% そこで，本研究では公平にする属性に対して仮定を置かず，個人間の公平性を考慮しながら，個人ごとに距離学習のマージンを設定する手法を提案する．
% そこで，本研究では属性ラベルフリーな学習手法を提案する
% This allows us to deal with biases in performance between different identities without making assumptions about which attributes should be fair.
% The proposed method uses individuals, the smallest unit of a group, to avoid the need for demographic group labeling that is typically required for fairness considerations. 
% This method utilizes the class confidence levels for each individual to introduce the concept of a fair class margin penalty based on the class favoritism level. 
% 提案手法は，属性ラベルを用いず(ラベルフリー)汎用的な学習手法でモデルを公平に学習することができる．
% また，提案手法を用いることで人種や他の属性に対しても公平性を向上させることを評価実験によって示す．
% Our proposed method allows for the learning of a model through a generic learning approach without using demographic annotations.
Comprehensive experiments have demonstrated that
our proposed method is effective for enhancing fairness while maintaining authentication accuracy.

% 公平な顔認証モデルを学習する手法として，多くの研究ではデータセットに付与された属性ラベルを用いている．
% しかしながら，顔画像データには照明条件や日焼けなどの様々な要因や，属性を正確に分類する定義が定まっているわけではないことから，属性を正確に分類することは難しい．
% そのため，顔画像に対してオートマティックに付与された属性ラベルを信用して公平な学習を行うことには問題がある．ddd
\end{abstract}
%TODO
\section{Introduction}
% 顔認証システムは生体認証システムの中でも，その利便性から近年急速に社会実装が進んだモダリティの1つである．機械学習システムのaccountabilityの議論が進む中で，顔認証システムもまた，その人種や性別といったでもグラフィック属性間の識別性能が一貫しないことが多く指摘されてきた\cite{biased}．
Face recognition is one of the modalities in biometric authentication systems that have seen rapid social adoption in recent years due to its convenience. As discussions about the responsibility of machine learning systems progress, it has often been pointed out that facial recognition systems also often show inconsistent performance in distinguishing between demographic attributes such as race and gender \cite{biased,jain}.

% ここいったん削除 %%%%%%%%%%%%%%%%%%%%%%%%%%%%%%%%%%%%%%%%%%%%%%%%%%%
% 顔認証システムの社会実装が進められていく中で，人種や性別等の属性間の公平性が指摘された．たとえば，Buolamwiniら\cite{biased}は，Microsoft，IBM，Face++等が構築した顔認識システムは白人男性の誤認識率が1\% 未満であるのに対して，黒人女性の誤認識率は35\% ほどであったことを報告している．実際に，米州に導入された犯罪捜査のための顔認証システムの誤認証は大半が黒人に対して発生しており，誤認逮捕に発展したケースが数多く報告されている\cite{investigate1, investigate2, investigate3}． % \cite{arrest1, arrest2, arrest3, arrest4}．このような公平性の指摘や誤認逮捕の社会問題を踏まえ，Amazonは警察組織に対する顔認証技術の提供を禁止\cite{amazon}し，IBMやMicrosoftが顔認証市場からの撤退を表明\cite{ibm, microsoft}するなど，大手テック企業が顔認証システムに対する慎重な動きを見せている．したがって，顔認証システムは利便性が高く様々な利用場面が想定される認証技術であるが，一部の利用場面において社会実装に対する社会的合意を満たすには，公平性課題の解決が必至である．
%  ここまで %%%%%%%%%%%%%%%%%%%%%%%%%%%%%%%%%%%%%%%%%%%%%%%%%%%

% 属性間の識別性能の偏りを解消するためのアルゴリズムは，大きく分類してデータセットの段階におけるアプローチとモデル学習段階におけるアプローチに大別される．データセットの構築段階では，人種割合のバランスが取れたデータセットを作成する\cite{bupt}，あるいは属性間の認識精度の偏りが小さくなるようなサンプリング，あるいはデータ拡張手法の提案\cite{class,mimick,imperfect}などが提案されてきた．モデル構築段階においては，人種に偏りのあるデータセットが存在するという仮定の下で，モデル性能のバイアスを取り除くために，属性間のスコア正規化や，ハイパーパラメータを属性に応じて動的に変動させる手法などが提案されてきた．
Approaches to mitigate bias in inter-attribute discrimination performance can be categorized into two stages: the dataset construction and the model construction.
In the dataset construction stage, efforts are made to create datasets with balanced racial proportions \cite{bupt, rfw}, sample or augment data to minimize disparities in recognition accuracy between attributes \cite{class,mimick}, and propose methods for data augmentation \cite{imperfect}.
In the model construction stage, strategies involve mitigating bias in model performance through score normalization between attributes \cite{score} and dynamically adjusting hyperparameters based on attributes \cite{bupt, labeling_dataset, mixfairface}, under the assumption of the existence of racially biased datasets.
% Algorithms for eliminating bias in inter-attribute discrimination performance can be broadly categorised into approaches at the dataset stage and those at the model learning stage.
% At the dataset construction stage, proposals have been made to create datasets with balanced racial proportions\cite{bupt, rfw}, to sample or augment data in ways that minimise disparities in recognition accuracy between attributes\cite{class,mimick}, and to suggest methods of data augmentation\cite{imperfect}.
% At the model construction stage, assuming that racially biased datasets exist, methods have been proposed to remove bias in model performance through score normalisation between attributes\cite{score} and dynamic adjustment of hyperparameters according to attributes\cite{bupt}.
%---------------------------------
% Most of previous approaches, whether dataset-based or model-based, have relied on methods assuming demographic group labeling or demographic classifiers, which cannot guarantee recognition accuracy for unrecognized attributes. 

Most previous approaches require sensitive attribute labels (e.g., race and gender) for training the network, which limits scalability to large-scale datasets and cannot guarantee accuracy for unknown attributes. This dependence on human-annotated labels poses challenges in terms of time, cost, and potential biases, especially for emerging attributes.
This paper proposes \textit{LabellessFace}, a novel framework that improves demographic bias in face recognition without requiring demographic group labeling. Our approach aims to maintain authentication accuracy while enhancing fairness. We introduce two key concepts: the \textit{class favoritism level}, quantifying the degree of favoritism towards specific classes across the dataset, and the \textit{fair class margin penalty}, extending existing metric learning methods based on class favoritism level.
% To address these problem, this paper proposes a novel framework, \textbf{LabellessFace}, which improves the demographic bias in face recognition without the need for demographic group labeling that is typically required for fairness considerations. 
% For this purpose, we introduce a novel fairness enhancement metric, class favoritism level, that considers the degree of favoritism towards specific class across the entire dataset. Based on the class favoritism level, we introduce the concept of a fair class margin penalty extending existing margin-based metric learning. This method dynamically incorporates class favoritism levels into learning parameters and improves overall fairness, i.e., it maintains fairness across all attributes. 
%
% By treating each class as an individual in facial recognition systems, we facilitate learning that minimizes biases in authentication accuracy among individuals. 
LabellessFace equalizes authentication accuracy across individuals without assuming specific sensitive attributes, achieving fairness even for unknown attributes.
We conducted comprehensive experiments using common facial benchmarks, demonstrating that our method successfully improves fairness while maintaining authentication accuracy comparable to existing approaches. The results show the effectiveness of LabellessFace in achieving fairness across both known and unknown demographic attributes.
Our contributions are summarized as follows:
\begin{itemize}
    \item We propose the concept of class favoritism levels, which quantifies the degree of favoritism towards specific class across the entire dataset.
    \item We propose the fair class margin penalty, which extends existing metric learning methods based on class favoritism levels. This realizes the LabellessFace framework that improves fairness without the need for labelling based on assumed target attributes.
    \item Comprehensive experiments have demonstrated that our proposed method is effective for enhancing fairness while maintaining authentication accuracy.
\end{itemize}

\section{Related Work}
\subsection{Fairness of Facial Recognition}
%顔認識システムは監視カメラの活用に伴い犯罪捜査等の重要な場面で用いられることもあり，人種に対する公平性の重要性は高く，多くの研究が行われている．
Facial recognition systems, increasingly integrated with surveillance cameras, are being deployed in critical scenarios such as criminal investigations, where the importance of racial fairness has been emphasised and numerous studies have been conducted.
%
%Buolamwiniら\cite{biased}は，Microsoft，IBM，Face++等の顔認識システムは白人男性の誤認識率が1\% 未満であるのに対し，黒人女性の誤認識率は35\% ほどであったことを報告している．
Buolamwini et al. \cite{biased} reported that facial recognition systems from companies such as Microsoft, IBM and Face++ had a misidentification rate of less than 1\% for white males, while the rate for black females was around 35\%.
%米国国立標準技術研究所（US National Institute of Standards and Technology, NIST）は，189の顔認証ソフトウェアを対象に人種に対する公平性の調査を行い，1：1認証において白人と黒人の誤検出率に10倍から100倍ほどの差が存在していたことや，1：N認証において黒人女性の誤検知率が高いことを示した\cite{nist}．
The US National Institute of Standards and Technology (NIST) conducted a fairness study of 189 facial recognition software systems. They found disparities in false positive rates between whites and blacks, with differences ranging from ten to a hundred times in 1:1 authentication scenarios. In addition, in 1:N authentication scenarios, black women had higher false detection rates \cite{nist}.
%
%Garvieら\cite{reason}は，学習データの人種割合の偏りが人種バイアスに大きな影響を与えていることを指摘し，エンジニアは白人の割合が高いことから意図せずに人間による差別が介入している可能性や，肌の色がコントラストに影響を与えること，女性の化粧が認証精度に影響を与える可能性があることを指摘した．
Garvie et al. \cite{reason} pointed out that the bias in the racial proportions of the training data significantly affects the racial bias. They highlighted the inadvertent introduction of human discrimination due to a higher proportion of Caucasians in the data, the effect of skin colour on contrast, and the potential effect of female make-up on authentication accuracy.

\subsection{Dataset-Based Approach}
% 機械学習の学習過程において，データセットの収集には人の手が加わるため，無意識のうちにバイアスが発生してしまう場合がある．そのため，顔認証システムにおいても人種バイアスの問題で最も着目されたのは学習データセットの人種割合である．
In the process of machine learning, the collection of datasets involves human intervention, which can unconsciously introduce bias. Therefore, in facial recognition systems, the issue of racial bias has been mainly related to the racial proportions in the training datasets.
% Wangら\cite{bupt}は，人種割合のバランスが取れたBUPT-Balancedfaceデータセットを作成し，学習を行うことで従来の人種割合が偏ったデータセットで学習を行った場合と比較して人種バイアスが軽減されることを示した．
Wang et al. \cite{bupt} created the BUPT-Balancedface dataset with balanced racial proportions and demonstrated that training with it could reduce racial bias compared to training with traditionally biased datasets in terms of racial proportions.
%
% Faisalら\cite{class}は，人種間の認識精度の偏りが小さくなるようにデータオブジェクトの置換を繰り返すリサンプリングを行うことで，人種バイアスを引き起こすデータをデータセットから取り除く手法を提案した．
Faisal et al. \cite{class} proposed a method of resampling in which data objects are repeatedly replaced to minimise differences in recognition accuracy between races, thereby removing data that could cause racial bias from the dataset.
%
% Qraitemら\cite{mimick}は，データセットの属性ごとの偏りを模倣した複数のサブセットを作成することで公平性を改善する手法を提案した．
Qraitem et al. \cite{mimick} proposed a method to improve fairness by creating multiple subsets that mimic the bias in the attributes of the dataset.
%
% また，学習データの収集において，DataAugmentation等の手法も活用されることがあるが，Niharikaら\cite{imperfect}はGANによるDataAugmentationによって人種バイアスを軽減させる際の性能的な限界を指摘している．
While techniques such as Data Augmentation are often employed in the collection of training data, Niharika et al. \cite{imperfect} have pointed out the performance limitations of reducing racial bias through Data Augmentation using GANs.
As an approach to optimizing a score threshold for a dataset, Pereira et al. \cite{marcel} introduced the Fairness Discrepancy Rate (FDR) to assess demographic differences by assuming a single decision threshold in biometric verification systems.

\subsection{Model-Based Approach}
% データセットの構築は人種間公平性を向上するための効率的な方法だが，公平性を考慮した大規模データセットは容易でない．これに加え，人種間の境界は曖昧であること\cite{palate}や，公平性の影響は人種と環境要因の相互関係によっても生じうること\cite{Sato2022}などが指摘されており，人種に偏りのある既存のデータセットにおいても公平に学習できるように，モデルの構造を工夫することで人種バイアス軽減に取り組む研究も行われてきた．
Dataset approach is an efficient method for improving fairness among races, yet constructing large-scale datasets with fairness considerations is not straightforward. Additionally, it has been pointed out that racial boundaries are ambiguous \cite{palate} and that the impact of fairness can also arise from the interrelationship between racial and environmental factors \cite{fairsa, Sato2022}.
Many researches has also been conducted to address racial bias by innovating the structure of models to allow fair learning even with existing datasets that are biased towards certain races. Most of the state-of-the-art methods and our proposed LabellessFace lies in this category.
% Philippら\cite{score}は認証対象の人種を事前情報とし，人種間のスコアを正規化するアプローチを用いている．
Philipp et al. \cite{score} used an approach that normalizes scores between races, using the race of the authentication subject as prior information.
%
% また，Dooleyら\cite{palate}は複数のモデルから公平性と精度のパレート最適解となるモデルを探索するアプローチを提案した．
Dooley et al. \cite{palate} proposed an approach to search for models that represent the Pareto optimal solution in terms of fairness and accuracy among multiple models.
%
% Wangら\cite{bupt}はArcface等の損失関数に含まれるハイパーパラメータを人種に応じて動的に変化させることで，人種割合が偏ったデータセットを用いて学習を行った場合でも人種間の認証精度の偏りを小さくできることを示した．
Xu et al. \cite{labeling_dataset} proposed an approach where they dynamically balances the false positive rate (FPR) between each training sample without the need for demographic labels.
Wang et al. \cite{mixfairface} proposed MixFair Adapter to estimate and reduce the identity bias, the performance inconsistency between different identities, by reducing the feature discriminability differences. 

% 我々が提案するLabelllessFaceは，損失関数に含まれるパラメータを動的に変化させる点でXu\cite{labeling_dataset}やWang\cite{mixfairface}の手法にインスパイアされた手法と言える．一方，Wang\cite{mixfairface}のように人工統計学的ラベルを必要とせず，Xu\cite{labeling_dataset}と異なりデータセット全体における特定クラスの優遇度合いを考慮した公平性向上手法を提案する．
Our proposed LabellessFace is inspired by the techniques of Xu et al. \cite{labeling_dataset} and Wang et al. \cite{mixfairface}, particularly because it does not require demographic labels. Unlike their approaches, we consider the degree of favoritism towards specific identity across the entire dataset.

\newcommand{\cossim}[2]{\cos\angle\left(#1,#2\right)}

\begin{figure*}
    \centering
    \includegraphics[width=0.9\textwidth]{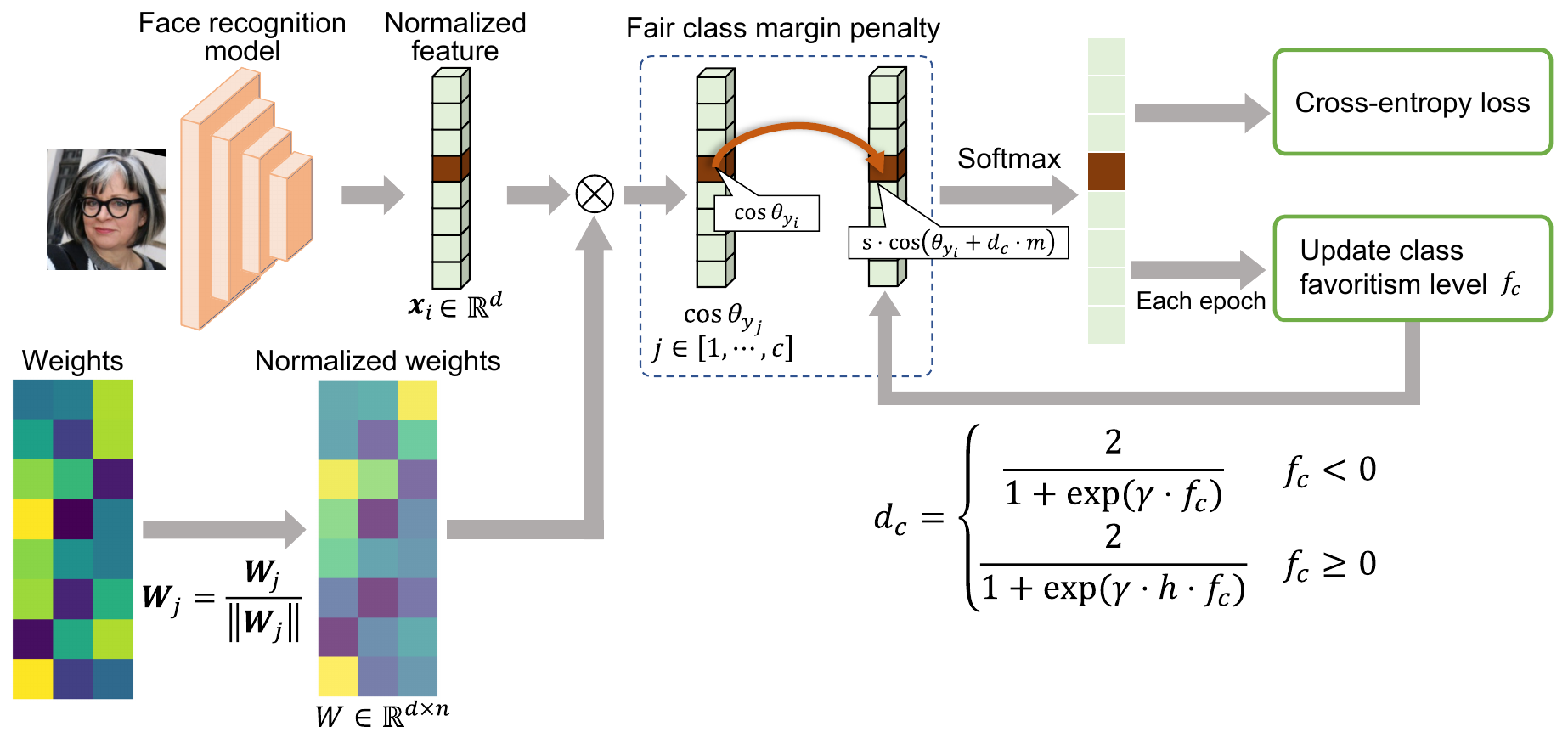}
    \caption{Overview of the LabellessFace framework.}
    \label{fig:overview_proposed}
\end{figure*}

\section{Proposed Method}
% 本節では，我々は Fair Class Margin Penalty を用いた LabellessFace framework の各コンポーネントについて紹介する．
In this section, we introduce each component of the \textit{LabellessFace} framework using the \textit{fair class margin penalty}.
% 本提案方式の概要を図\ref{fig:overview_proposed}に示す．既存のソフトマックスに基づく距離学習（\ref{subsec:softmax}節）に加え，提案手法は，クラス優遇度に基づき各クラスごとに異なるマージンを動的に設定しながら学習を進めるFair Class Margin Penaltyプロセス（\ref{subsec:fair_class_margin_penalty}節），および各エポックの終了時にクラス優遇度を算出するプロセス（\ref{}節）から構成される．クラス優遇度は各クラスの識別精度が全体平均からどの程度偏っているかを学習サンプルに基づいて決定する．本提案方式では，クラス優遇度をクラスの最小単位である個人に対して算出することで，学習データ属性に関するラベリングを不要にしつつ，
The overview of the proposed method is shown in Figure \ref{fig:overview_proposed}. In addition to existing softmax-based metric learning (section \ref{subsec:softmax}), our method dynamically sets different margins for each class based on \textit{class favoritism level} while progressing the training through the fair class margin penalty process (section \ref{subsec:fair_class_margin_penalty}), and updates the class favoritism level at the end of each epoch (section \ref{subsec:class_favoritism_level}). Here, the class favoritism level is determined based on how much the recognition accuracy for each individual deviates from the overall average using the training samples.

\subsection{Softmax-Based Metric Learning}
\label{subsec:softmax}
%オリジナルのソフトマックス損失関数は以下のように示される．
The original softmax loss function is formulated as follows: 
\begin{align}
\label{formula:Softmax-CrossEntropy-detail}
\mathcal{L} = -\log\frac {e^{\bm{W}_{y_i}^T\bm{x}_i+b_{y_i}}}{\sum_{j=1}^{|C|} e^{\bm{W}_j^T\bm{x}_i+b_j}},
\end{align}
%
% $\bm{x}_i \in \mathbb{R}^d $はクラス$y_i$に属する全結合層への入力となる$d$次元特徴ベクトルであり，$\bm{W}_j \in \mathbb{R}^d$は重み$W\in \mathbb{R}^{d\times |C|}$の$j$列目を表す．また，$b_j$はバイアス項を表す．このとき，バイアス項$b_j=0$とし，L2正則化により$||\bm{W}_j|| = 1, ||\bm{x}_i||=1$となる制約を行った場合，式（\ref{formula:Softmax-CrossEntropy-detail}）は$\bm{W}_{y_i}$と$x_i$のなす角を$\theta_{y_i}$とし，式（\ref{formula:Softmax-CrossEntropy-dot}）のように表される．
where $\bm{x}_i \in \mathbb{R}^d$ is a $d$-dimensional feature vector that serves as the input to the fully-connected layer corresponding to class $y_i$, and $\bm{W}_j \in \mathbb{R}^d$ represents the $j$-th column of the weight matrix $W \in \mathbb{R}^{d \times |C|}$. Additionally, $b_j$ represents the bias term. 
Under the conditions where the bias term $b_j=0$ and L2 regularization constraints $||\bm{W}_j|| = 1$ and $||\bm{x}_i|| = 1$ are applied, equation (\ref{formula:Softmax-CrossEntropy-detail}) expresses the angle between $\bm{W}_{y_i}$ and $x_i$ as $\theta{y_i}$, and can be represented by
\begin{align}
\label{formula:Softmax-CrossEntropy-dot}
\mathcal{L} = -\log\frac {e^{\cos\cdot{\theta_{y_i}}}}{e^{\cos\cdot{\theta_{y_i}}} + \sum_{j=1, j\neq y_i}^{|C|}e^{\cos\cdot{\theta_j}}}.
\end{align}
% 一般に，学習が困難なハードサンプルは$||\bm{x}_i||$が小さくなるように学習することが知られている\cite{smallnorm}．
% そのため，L2正則化により$||\bm{W}_j|| = 1, ||\bm{x}_i||=1$となる制約を行うことでハードサンプルの学習を正しく行うことが期待されている．
% そして，式（\ref{formula:Softmax-CrossEntropy-dot}）の正解クラス部を変調関数$T(\cdot)$, 不正解クラス部を$N(\cdot)$に置き換え，スケールパラメータ$s$を適用した損失関数$\mathcal{L}$は式（\ref{formula:replace}）のように表される．
%
% 削除05/01
% It is known that difficult-to-learn hard samples often result in small norms for $\bm{x}_i$ during training \cite{smallnorm}. Thus, by applying L2 regularization to enforce $||\bm{W}_j|| = 1$ and $||\bm{x}_i|| = 1$, it is expected that learning for hard samples will be correctly addressed. 
% With this, the correct class term of equation (\ref{formula:Softmax-CrossEntropy-dot}) is replaced by a modulation function $T(\cdot)$, and the incorrect class term by $N(\cdot)$, applying a scale parameter $s$. The loss function $\mathcal{L}$ is then represented by
%
% \begin{align}
% \label{formula:replace}
% \mathcal{L} = -\log\frac {e^{sT(\cos {\theta_{y_i}})}}{e^{sT(\cos {\theta_{y_i}})} + \sum_{j=1, j\neq y_i}^{|C|}e^{sN(t, \cos {\theta_j})}}.
% \end{align}
%
% ソフトマックス関数に基づく距離学習では，Angular Margin Penalty項を用いて正解クラス$y_i$に変調関数$T(\cdot)$を用いることでクラス内分散を小さくするようにし，スケールパラメータ$s$を用いて$\cos {\theta_i}$をスケールしている．変調関数$T(\cdot)$は各距離学習メトリクスによって異なる．
% 例えば，ArcFaceの場合，マージンパラメータ$m$を用いて変調関数$T(\cdot)$を式（\ref{formula:arcface}）のように定義している．
%
% ソフトマックス関数に基づく距離学習では，Angular Margin Penalty項を用いて正解クラス$y_i$のクラス内分散を小さくするようにし，スケールパラメータ$s$を用いて$\cos {\theta_i}$をスケールしている．例えば，ArcFaceの場合，マージンパラメータ$m$を用いて式\ref{formula:replace}を次式（\ref{formula:arcface}）のように定義している．
%
In metric learning based on the softmax loss, the Angular Margin Penalty is employed to reduce intra-class variance for the correct class $y_i$, and a scale parameter $s$ is used to scale $\cos{\theta_i}$. For example, in ArcFace \cite{arcface}, the equation is defined by adding the margin parameter $m$ to equation (\ref{formula:Softmax-CrossEntropy-dot}) as follows:
\begin{align}
\label{formula:arcface}
\mathcal{L} = -\log\frac {e^{s\cdot(\cos{\theta_{y_i}+m})}}{e^{s\cdot(\cos {\theta_{y_i}+m})} + \sum_{j=1, j\neq y_i}^{|C|}e^{s\cdot(\cos {\theta_j})}}.
\end{align}
%
% 本提案手法では，式（\ref{formula:arcface}）に対して，クラスごとに最適なマージンが動的に変化するような係数$d_c$を追加する．
In the proposed method, a coefficient is added to the equation (\ref{formula:arcface}) that allows the optimal margin to vary dynamically for each class.

\subsection{Fair Class Margin Penalty}
\label{subsec:fair_class_margin_penalty}
\begin{figure*}
    \centering
    \includegraphics[width=0.9\textwidth]{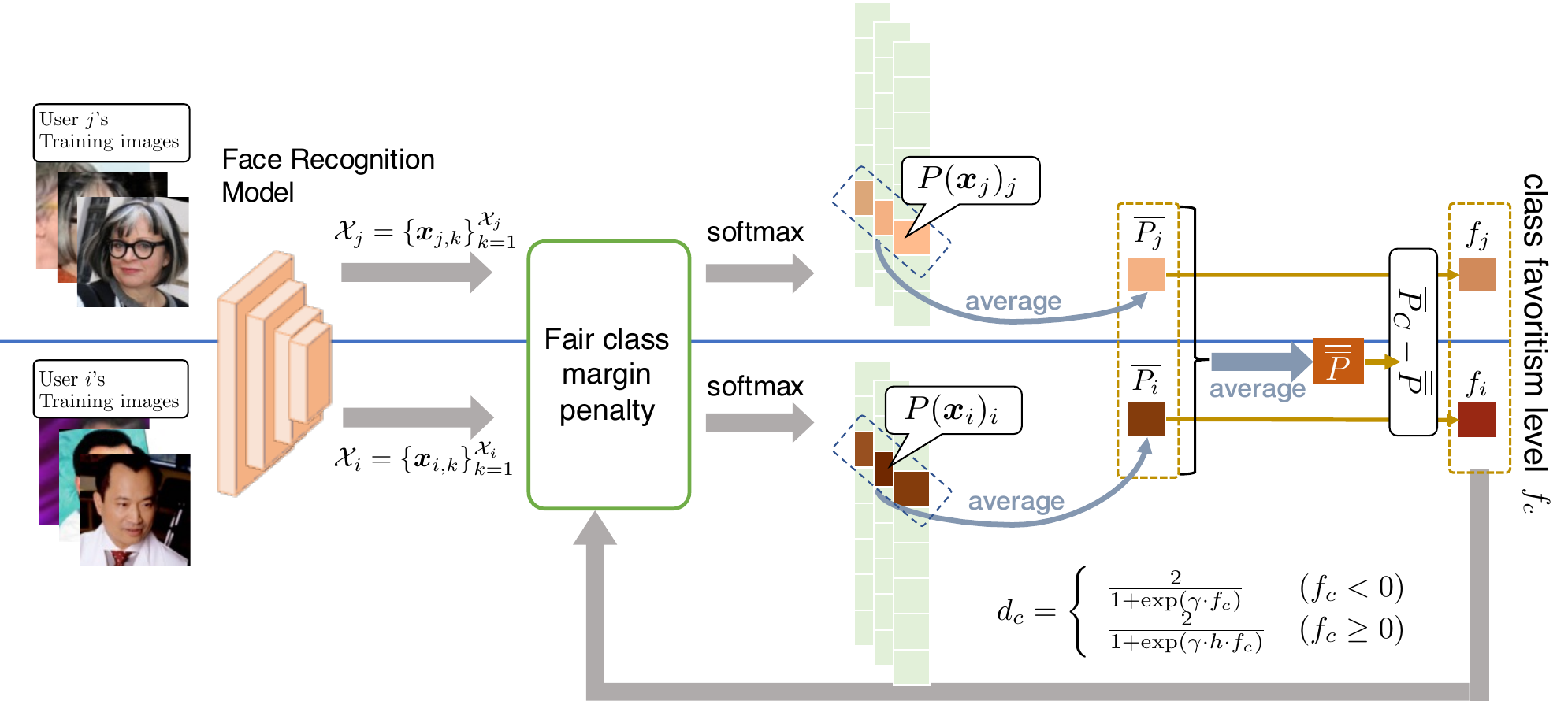}%{fig/overview_cfl.pdf}%
    
    \caption{Overview of the Class Favoritism Level calculation.}
    \label{fig:overview_cfl}
\end{figure*}
%
% 本提案では，個人ごとの認証精度の偏りを最小にするために，式（\ref{formula:arcface}）に対して，クラスごとに最適なマージンが学習中に動的に変化するような係数$d_c$（以下，マージン係数）を追加する．これにより，式（\ref{formula:arcface}）で定義した変調関数$T_c$を式（\ref{formula:original_T}）とする．
In this proposal, to minimize the bias in individual authentication accuracy, a coefficient $d_c$ (hereafter referred to as the margin coefficient) is added to equation (\ref{formula:arcface}) so that the optimal margin for each class dynamically changes during training. Consequently, equation (\ref{formula:arcface}) is modified as
% %
% \begin{align}
% \label{formula:original_T}
% T_c(\cos \theta_{y_i}) = \cos (\theta_{y_i} + d_c \cdot m)
% \end{align}
% %
\begin{align}
\label{formula:original_T}
\mathcal{L} = -\log\frac {e^{s(\cos{\theta_{y_i}+d_c \cdot m})}}{e^{s(\cos {\theta_{y_i}+d_c \cdot m})} + \sum_{j=1, j\neq y_i}^{|C|}e^{s\cdot(\cos {\theta_j})}}.
\end{align}
%
%
% ここで，$d_c$は，クラスごとに異なる値となり，各エポックの終了時に各クラス$c \in \mathcal{C}$が全クラスのうちどの程度優遇されているかを示すクラス優遇度$f_c$に基づき決定される．
% マージン係数$d_c$は式（\ref{formula:original_D}）のように定義される．
In this case, $d_c$ takes different values for each class and is determined at the end of each epoch based on the class favoritism level $f_c$, which indicates the extent to which each class $c \in \mathcal{C}$ is favored among all classes. The margin coefficient $d_c$ is defined by
\begin{align}
\label{formula:original_D}
d_c=\left\{ \begin{array}{ll} 
\frac{2}{1 + \exp(\gamma \cdot f_c)} & (f_c < 0) \\
\frac{2}{1 + \exp(\gamma \cdot h \cdot f_c)}& (f_c \geq 0) \end{array}\right..
\end{align}
% マージン係数$d_c$は，冷遇されているクラスに対してマージンの影響度を大きくすることで顔特徴領域を広く確保しようとし，優遇されているクラスに対してマージンの影響度を小さくすることで顔特徴領域を狭くして公平性を向上させようとする．
The margin coefficient $d_c$ is designed to increase the margin's impact on classes that are less favored, thereby enlarging the facial feature space, while reducing the margin's impact on classes that are more favored, thus narrowing the facial feature space to enhance fairness.

\begin{figure}[t]
 \begin{center}
 \includegraphics[scale=0.5]{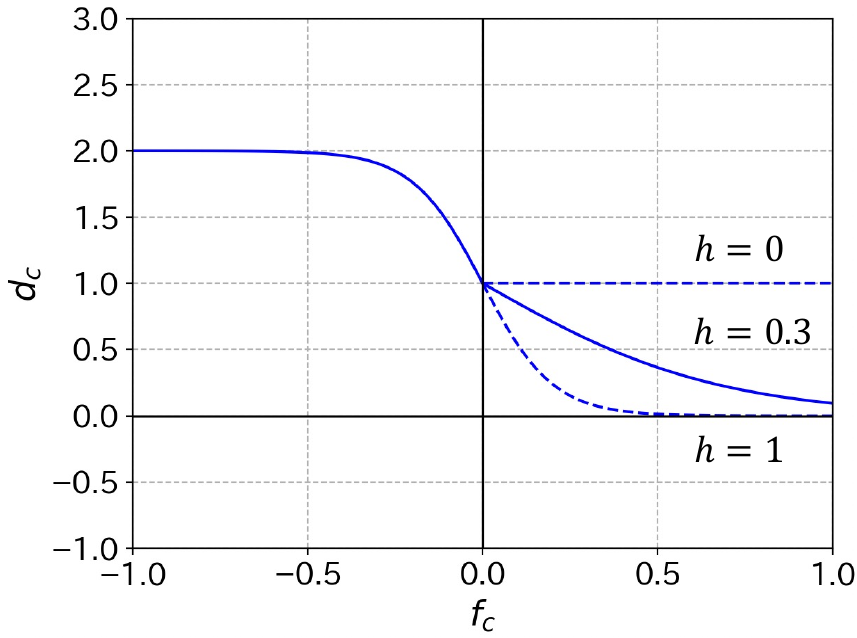}
 \caption{The gradient change of the margin coefficient $d_c$ with respect to the value of the harmony coefficient.}
 \label{fig:harmony}
 \end{center}
\end{figure}

% \noindent\textbf{Hyperparameters. }図\ref{fig:harmony}に$d_c$とクラス優遇度$f_c$との関係を示す．(\ref{formula:original_D})式における係数$\gamma$は$[0,\infty)$の実数を取り，マージン係数$d_c$の勾配を決定するハイパーパラメータである．勾配係数$\gamma=0$のとき，クラス優遇度$f_c$の値に依らずマージン係数$d_c=1$になり，ArcFaceと等しい．また，係数$h$は，$[0,1]$の実数を取り，クラス集合間の公平性の向上をどれほど重要視するかを決定するハイパーパラメータである．値が小さいほど優遇されているクラスの精度低下を許容せず，公平性よりも認証精度の重要度を重視する．また，値が大きいほど優遇されているクラスの精度低下を許容することで，認証精度よりも公平性の重要度を重視する．なお，本提案では$\gamma$を勾配係数，$h$を調和係数とそれぞれ呼ぶ．
\noindent\textbf{Hyperparameters. } Figure \ref{fig:harmony} shows the relationship between $d_c$ and the class favoritism level $f_c$. In equation (\ref{formula:original_D}), the coefficient $\gamma$ is a real number within the range $[0, \infty)$ and serves as a hyperparameter that determines the gradient of the margin coefficient $d_c$. When the gradient coefficient $\gamma=0$, the margin coefficient $d_c$ becomes 1, irrespective of the value of class favoritism level $f_c$, making it equivalent to ArcFace \cite{arcface}. Additionally, the coefficient $h$, taking a real number in the range $[0,1]$, determines how much importance is placed on improving fairness among class sets. A lower value of $h$ indicates a reluctance to allow a decrease in accuracy for favored classes, prioritizing authentication accuracy over fairness. Conversely, a higher value of $h$ allows for a decrease in accuracy of favored classes, emphasizing fairness over authentication accuracy. In this paper, $\gamma$ is referred to as the \textit{gradient coefficient} and $h$ as the \textit{harmony coefficient}.

\subsection{Class Favoritism Level Calculation}
\label{subsec:class_favoritism_level}
%クラス$c$に対するクラス優遇度$f_c$は各エポックの終了時に学習データを用いて算出され，次エポックにおけるマージン係数$d_c$の計算に反映される．
The class favoritism level $f_c$ for class $c$ is calculated at the end of each epoch using the training data and is reflected in the calculation of the margin coefficient $d_c$ for the subsequent epoch.

% ソフトマックスの出力から得られる確信度は学習が十分行われればサンプルが属するクラスの確信度が最大となり，それ以外のクラスの確信度は小さくなるが，この傾向はサンプルが属するクラスに依存して異なる．このとき，各クラスが取りうる確信度に対して相対的に高い確信度で判別されるクラスは，クラス集合において優遇され，一方，低い確信度で判別されるクラスは冷遇されていると捉えることができる．
The confidence derived from the softmax output increases for the class to which a sample belongs as training progresses, maximizing the confidence for the sample's class while reducing the confidence for other classes. However, this tendency varies depending on the class to which the sample belongs. At this point, classes identified with relatively high confidence within the class set can be considered as favoured, whereas classes identified with low confidence are perceived as neglected.
% クラス優遇度は$f_c$は，このようなクラス間の確信度の差を公平性指標として捉え，定量化する．クラス優遇度$f_c$の算出方法の概要を図\ref{fig:overview_cfl}に示す．
The class favoritism level $f_c$ interprets and quantifies the difference in confidence levels among classes as a measure of fairness. An overview of the method for calculating the class favoritism level $f_c$ is shown in Figure \ref{fig:overview_cfl}.

% クラス$c$の学習データから取得した特徴量を$\mathcal{X}_c=\{\bm{x}_{c,k}\}$ $(k=1,\cdots,|\mathcal{X}_c|)$ とする． ここで，$|\mathcal{X}_c|$はクラス$c$の全学習データ数を示す．
% $\bm{x}_{c,k}$を顔認証モデルに入力して，特徴量に対するsoftmax出力$P(\bm{x}_{c,k})$を得る．ここで，$P(\bm{x}_{c,k})$のうち，クラス$c$に対応する成分の確信度を$P(\bm{x}_{c,k})_c$とすれば，$\bm{x_{c,k}}$に対する確信度の平均$\overline{P_c}$を式（\ref{formula:p_bar}）で示すことができる．
Let the features extracted from the training data of class $c$ be denoted as $\mathcal{X}_c = \{\bm{x}_{c,k}\}$ $(k=1,\cdots,|\mathcal{X}_c|)$, where $|\mathcal{X}_c|$ represents the total number of training data for class $c$. Inputting $\bm{x}_{c,k}$ into the face recognition model yields the softmax output $P(\bm{x}_{c,k})$. Here, denoting the confidence component corresponding to class $c$ in $P(\bm{x}_{c,k})$ by $P(\bm{x}_{c,k})_c$, then the average confidence for $\bm{x}_{c,k}$ can be expressed as $\overline{P_c}$, which is given by
%
%を式（\ref{formula:p_cx}）に示す．
% ここで，$\varpi_{c}$はsoftmaxから出力される確信度の$c \in C$成分を返す関数である．
% \begin{equation}
% \label{formula:p_cx}
% P(c,x) = \varpi_{c}(\text{softmax}(Z(x)))
% \end{equation}
%
\begin{equation}
\label{formula:p_bar}
\overline{P_c} = \operatorname{Mean}(P(\bm{x_}{c,k})_c)=\frac{1}{|\mathcal{X}_c|}\sum_{k=1}^{|\mathcal{X}_c|} P(\bm{x}_{c, k})_c.
\end{equation}
%
% さらに，全てのクラス$c \in \mathcal{C}$について$\overline{P_c}$を求めたのち，これらの平均$\overline{\overline{P}}$を式（\ref{formula:p_barbar}）によって求める．
After calculating $\overline{P_c}$ for all classes $c \in \mathcal{C}$, the average of these values, $\overline{\overline{P}}$, is derived as follows:
\begin{equation}
\label{formula:p_barbar}
\overline{\overline{P}} = \frac{1}{|C|}\sum_{c \in C} \overline{P_c}.
\end{equation}
%
%クラス$c$に対するクラス優遇度$f_c$は，クラス集合$C$におけるクラス$c$の相対的な優遇度として計算できる．したがって，各クラスの$\overline{P_c}$と$\overline{\overline{P}}$の差分により，式（\ref{formula:original_f_c}）のように定義する．
The class favoritism level $f_c$ for class $c$ can be derived as the relative favoritism of class $c$ within the class set $\mathcal{C}$. Therefore, the class favoritism level $f_c$ is defined by the difference between $\overline{P_c}$ for each class and the average $\overline{\overline{P}}$ as follows:
\begin{equation}
\label{formula:original_f_c}
f_c = \overline{P_c} - \overline{\overline{P}},
\end{equation}
% ここで，$\overline{P_c}$および，$\overline{\overline{P}}$はそれぞれ[0,1]の範囲を取り，クラス優遇度$f_c$は[-1,1]の範囲を取る．例えば，クラス優遇度が負の値を取るクラスは，そのクラスに属するサンプルが，相対的に低い確信度で識別が行われるため，認証を誤る可能性が他クラスよりも高く，該当するクラスは冷遇されていると捉えることができる．クラス優遇度$f_c$は各エポック終了時に評価されて更新される．したがって，動的マージン係数は，学習過程において定期的にクラス優遇度$f_c$を参照することで，それぞれのクラスが優遇されているのか，冷遇されているのかを監視し，それぞれのクラスに合った最適なマージンをエポックごとに再設定する．
where the range of $\overline{P_c}$ and $\overline{\overline{P}}$ is [0,1], and the range of $f_c$ is [-1,1]. For instance, a class with a negative class favoritism level $f_c$ indicates that samples belonging to that class are identified with relatively low confidence, thus they are more prone to misclassification compared to other classes, and such classes can be considered as neglected. Therefore, the dynamic margin coefficient $d_c$ monitors whether each class is favored or neglected by referencing the class favoritism level $f_c$ regularly throughout the training process, and adjusts the optimal margin for each class at every epoch accordingly.

% Ratial 使わないので削除 --- 2024/04/30
% なお，本提案手法において，クラス$c$は人種から個人の単位まで柔軟に設定が可能である．人種ごとにクラス優遇度を算出することで人種間の公平性を考慮することができる．本提案では，クラス$c$を人種単位としたモデルを\texttt{Ratial}，これに対して，クラス$c$を個人単位としたモデルを\texttt{Identity}と呼ぶ．

% % ←→ proposed: Class Margin Method
% \subsection{Attribute Margin Method}

% 本実験では，クラス単位のマージン変化を行う本提案手法を比較評価するため，属性単位でマージン変化させた属性考慮モデルを作成する．

% 特に，学習データセットに付与された人種ラベルを用いて学習を行ったモデルを，\textbf{人種考慮モデル}と表記し，本実験ではAfrican，Asian，Caucasian，Indianの4人種の事前にラベリングされた情報を参照して学習を行う．
\section{Experiments}
% 5. 実験の概要説明←実験の目的について書く
% 5.1 Protocol
% 5.1.1 Dataset←利用したDBとtrain/testの分割などについてまとめて書く
% 5.1.2 Implementation←提案手法の実装の詳細，エポック数その他について書く
% 5.2 Result←これから紹介する実験結果について補足する．たとえば比較したモデルなど．
% ArcFaceと自分の手法と比較した．それぞれArcFace/Racial/Proposedと書くなど．
% 5.2.1 実験1の結果について
% 5.2.2 実験2の結果について
% 5.2.3 ….
%
% まず，本実験で用いる識別性能や公平性について評価するための指標の定義を説明し，本実験の研究課題とそれに向けた実験の簡単な流れを説明した後，実験の実装詳細と結果を示す．
% \subsection{Scenario}
% 本実験では，以下の研究課題を明らかにすることでクラスごとに最適なマージンを動的に設定するクラスマージン手法の有効性を評価する．

% \begin{itemize}
%     \item クラスマージン手法は他手法と比較して認証精度はどのように変化するか
%     \item クラスマージン手法は他手法と比較して人種間の公平性を向上させるか
%     \item クラスマージン手法は人種以外の様々な未知属性間の公平性を向上させるか
% \end{itemize}

% それぞれの研究課題を明らかにするための実験概要を述べる．
% First, we define the metrics used to evaluate discriminative performance and fairness in this experiment. Following this, we describe the research questions addressed by this experiment and outline the experimental flow. Finally, we present the implementation details and results of the experiment.
In our experiment, we evaluate the effectiveness of a our proposed Labelless Face framework by addressing the following two viewpoints: ``\textbf{(1)} Can our proposed method improve the fairness of certain sensitive attributes (e.g. race or gender) while maintaining accuracy? (Section \ref{subsec:eval_performance})'' and 
``(2) Can our proposed method improve fairness independent of annotated labels? (Section \ref{subsec:eval_fairness})''.
% In our experiment, we evaluate the effectiveness of a our proposed Labelless Face framework by addressing the following three viewpoints, \red{\textbf{(1)} Can our proposed method improve the fairness of certain sensitive attributes (e.g. race or gender) while maintaining accuracy? (Section \ref{subsec:eval_performance})}
% \red{(2) Can our proposed method improve fairness independent of annotated labels? (Section \ref{subsec:eval_fairness})}
We describe the experimental protocols used to address each research question, present the results, and discuss their implications.

\subsection{Protocol}
% 実験環境を表\ref{env}に示す．
\begin{table}[t]
  \caption{Experimental environments.}
  \label{env}
  \centering
  \scalebox{1.0}{
  \begin{tabular}{ll}
    \hline \hline
    Language & Python3.8.10\\
    GPU & NVIDIA RTX 6000 Ada $\times$ 4\\
    CPU & Intel(R) Xeon(R) Gold 5418Y\\
    Memory & 256GB\\
    Kernel & 6.2.0-32-generic\\
    Distribution & Ubuntu 22.04.3 LT\\
    \hline
  \end{tabular}
  }
\end{table}

\noindent\textbf{Training Dataset. }\label{dataset}
% モデルの学習には人種割合が統一されたBUPT-Balancedface\cite{bupt}を用いた．
% BUPT-BalancedfaceはAfrican, Asian, Caucasian, Indianの4人種が各7,000クラス含まれたデータセットであり，各データに人種のラベリングがされている．4人種計28,000クラスの顔画像を，学習と検証の割合が9：1になるように分割し，検証のデータは早期学習終了の判定とクラス優遇度$\mathbf{f}_c$を算出するために用いる．
For model training, the BUPT-Balancedface dataset \cite{bupt}, which has an equal proportion of races, was used. BUPT-Balancedface contains 7,000 classes of four races: African, Asian, Caucasian, and Indian, with racial labeling provided for each data point. The dataset, comprising 28,000 classes of facial images from these four races, was split into a training and validation ratio of 9:1. The validation data is used for early stopping decisions and for calculating the class favoritism levels $f_c$.

% 評価では，同一人物・他人の顔画像ペアが用意されたLFW\cite{lfw}と4人種のラベリングがされたRFW\cite{rfw}を用いた．LFWは事前に同一人物ペア（genuine），他人ペア（imposter）のそれぞれ約3,000組の顔画像ペアセットが用意されたデータセットであり，各クラスの人種や年齢，髪色や眼鏡の有無といった計74属性がラベル付けされており，モデルの識別性能評価や，それぞれの属性ドメインの公平性評価を行うために用いた．

% RFWは，BUPT-Balancedfaceと同様にAfrican, Asian,Caucasian, Indianの4人種がそれぞれ約3,000クラス含まれたデータセットであり，各データに人種のラベリングがされている．RFWの同一人物ペア，他人ペアのそれぞれ約3,000組の顔画像ペアセットを作成し，1:1認証における公平性評価を行なうために用いた．
\noindent\textbf{Evaluation Datasets. }
In the evaluation, the Labeled Faces in the Wild (LFW) \cite{lfw} and Racial Faces in the Wild (RFW) \cite{rfw} datasets were used. LFW is a dataset with approximately 3,000 pairs each of genuine and imposter facial image pairs prepared in advance, and each class is labeled with 74 attributes including race, age, hair color, and the presence of glasses. This dataset is used for evaluating the discriminative performance of the model and assessing fairness across various attribute domains. RFW contains about 3,000 classes for each of the four races: African, Asian, Caucasian, and Indian, with racial labeling provided for each data point. Approximately 3,000 pairs each of genuine and imposter facial image pairs were created using RFW, and it was used to evaluate fairness in 1:1 verification.

\noindent\textbf{Implementation Detail. }
% モデルはResNet34を使用し，最終層手前の層を距離学習レイヤと接続し28,000クラスの分類器として学習を行った．ResNet34のパラメータはランダムに初期化し，学習時のバッチサイズは256，学習率は1e-1から1e-4まで線形に変化させた．weight decayは5e-5，モーメンタムは0.9とした．また，最適化アルゴリズムにはSGDを用いて30エポックの学習を行った．距離学習のスケール$s$は64，初期マージン$m$は0.3とし，提案手法のハイパーパラメータである勾配係数$\gamma$と調和パラメータ$h$は本実験では$\gamma=10, h=1$とした．学習後，1：1認証シナリオにおいて特徴抽出器として使用する場合は512次元の最終層手前の出力を特徴ベクトルとして用いた．
We utilized ResNet34 \cite{He-CVPR-2016} as a face recognition model architecture, with the layer prior to the final layer connected to a metric learning layer, trained as a classifier for 28,000 classes. The parameters of ResNet34 were initialized randomly. The batch size during training was set to 256, and the learning rate was linearly adjusted from $1e-1$ to $1e-4$. The weight decay was set at $5e-5$ and momentum at $0.9$. The optimization algorithm used was SGD, and training was conducted over 30 epochs. For metric learning, the scale $s$ was set to 64, and the initial margin $m$ was set to $0.3$. The hyperparameters for the proposed method were set to the gradient coefficient $\gamma=10$ and the harmony coefficient $h=1$ for the experiments. 
%After training, for the 1:1 verification scenario, the output from the layer prior to the final layer, which is 512-dimensional, was used as the feature vector.

% \noindent \textbf{Performance Experiment. }% TODO: MagFaceの説明は3章(Technical Background)でするべきか？
% 距離学習手法であるArcFaceとMagFace\cite{magface},人種考慮モデル，そして提案手法モデルの三つのモデルに対して認証精度の評価を行う．

% 学習データセットには，African，Asian，Caucasian，Indianの4人種がラベリングされ各7,000クラス含まれたBUPT-Balancedface\cite{bupt}をすべての実験で共通して用いる．

% テストデータセットには，顔画像ペアの組が事前に提供されているLFW\cite{lfw}を用いて認証精度の評価を行う．

% LFWの顔画像ペアに対して1：1認証を行い，性能評価指標のEER， AUC，ROC曲線，DET曲線を算出して各モデルの比較評価を行う．

% \noindent \textbf{Racial Fairness Experiment}

% 距離学習手法であるArcFaceとMagFace,人種考慮モデル，そして提案手法モデルの三つのモデルに対して人種間の公平性の評価を行う．

% テストデータセットには，BUPT-Balancedfaceと同じ方法で人種がラベリングされているRFW\cite{rfw}を用いる．

% BUPT-Balancedface，RFWはとも  にMS-Celeb-1M\cite{msceleb}の顔画像に対してFreeBase\cite{FreeBase}で国籍情報を参照することでAsian，Indianを収集し，Free++\cite{facepp} APIを用いてCaucasian，Africanの人種推定を行い人種をラベリングしている．

% また，BUPT-Balancedfaceを構築する際には，RFWと重複のないようにスクリーニングされているため，BUPT-Balancedfaceを学習データに，RFWをテストデータに使用することができる．

% 各データセットは同じ方法で人種をラベリングしていることから，人種ラベルを参照して人種ごとにマージンを動的に設定して学習した人種考慮モデルは，テストデータセットの属性ドメインを考慮して学習されたモデルであるといえる．

% RFWの顔画像ペアに対して1：1認証を行い，公平性評価指標のSTD，Gini，SERを算出して各モデルの比較評価を行う．

% \noindent \textbf{Other Attributes Fairness Experiment}

% 距離学習手法であるArcFaceとMagFace,人種考慮モデル，そして提案手法モデルの三つのモデルに対して様々な未知属性間の公平性の評価を行う．

% テストデータには，計74の属性がラベリングされたLFWを用いる．

% サンプル数の偏りを考慮し，実験詳細（\ref{dataset}）で後述する方法を用いてスクリーニングした26属性のうち，テンプレート画像，照合画像が含んでいる属性間のEERを計算することで，公平性評価指標のSTD，Gini，STDを算出して各モデルの比較評価を行う．

\subsection{Metrics}
%モデルの識別性能については，Equal Error Rate（EER）とArea Under Curve（AUC）による定量的な評価を行う．
We quantitatively evaluated the discriminative performance of the proposed method by the Equal Error Rate (EER) and the Area Under the Curve (AUC). 
% と，ROC曲線，DET曲線による定性的な評価を行う．
% EERは，False Non-Match Rate（FNMR）とFalse Match Rate（FMR）が等しくなるように，入力された画像ペアが本人であるか他人であるかを決定するしきい値を事後的に調整した際のエラー率であり，低いほど識別性能が高いといえる．
% FNMRは本人を誤って拒否してしまう率を指し，FMRは他人を誤って受け入れてしまう率を指す．
% ROC曲線とは，本人であるか他人であるかを決定するしきい値を変動させた際のFNMR，FMRの変動を二軸にプロットした曲線である．FNMRとFMRのトレードオフ性を示し，ROC曲線の下側面積がAUCであり，高いほど識別性能が高いといえる．
% DET曲線は，定性的な評価を行いやすいように，モデルの1-FNMRとFMRの変動を正規分布軸にプロットした曲線であり，モデルの性能が直線的に表現される．原点との距離が近いほど識別性能が高いといえる．
%
%モデルの公平性については，各属性間のEERの標準偏差（STD），Gini Index（Gini），Skewed Error Ratio（SER）で評価する．
%ジニ係数\cite{gini}は各属性間のEER累積分配比率の偏りを表す公平性指標であり，式（\ref{formula:gini}）のように定義される．
%SERはエラー率の最も高い属性と最も低い属性の比を表す公平性指標であり，式（\ref{formula:ser}）のように定義される．
%ここで，$n$は属性の総数，$EER_i, EER_j$は属性$i, j$の$EER$を指す．
The fairness of the model is evaluated by the standard deviation of EER (STD), the Gini Index (Gini), and the Skewed Error Ratio (SER) across different classes. The Gini coefficient \cite{gini} is a fairness metric that represents the disparity in the cumulative distribution ratios of EER among different classes, and is defined by
\begin{align}
  \label{formula:gini}
  \mathrm{Gini} = \frac{\sum_{i=1}^{|\mathcal{C}|}\sum_{j=1}^{|\mathcal{C}|}|EER_i - EER_j|}{2|\mathcal{C}|^2\overline{EER}},
\end{align}
where $|\mathcal{C}|$ represents the total number of classes, and $EER_i$, $EER_j$ refer to EER for classes $i$ and $j$, respectively, $\overline{EER}$ represents the average of EERs.
SER is a fairness metric that represents the ratio between the highest and lowest error rates among attributes, and is defined by
\begin{align}
  \label{formula:ser}
  \mathrm{SER} = \frac{\max_{c\in\mathcal{C}}(EER_c)}{\min_{c\in\mathcal{C}}(EER_c)}.
\end{align}

\begin{table*}[t]
 \setlength{\captionmargin}{50pt}
 \caption{The performance and fairness evaluation results trained on BUPT-Balancedface dataset and evaluated on RFW dataset: Ratial attributes (Asian, African, Caucasian, Indian, referred to as As, Af, Ca, In, respectively) were selected as the subjects for evaluation. STD, Gini, SER were assessed when users were divided according to these attributes, respectively.}
 \label{table:performance}
 \centering
  %\begin{tabular}{c|cc|cccc|ccc}
 \begin{tabular}{cccccccc}
   \toprule
   & EER-Af($\downarrow$) & EER-As($\downarrow$) & EER-Ca($\downarrow$) & EER-In($\downarrow$) & STD($\downarrow$) & Gini($\downarrow$) & SER($\downarrow$) \\
   \midrule
   \texttt{ArcFace} & 0.1847 & 0.1975 & 0.1145 & 0.1621 & 0.03163 & 0.01031 & 0.1725 \\%& 0.03163 & 0.1031 & 1.725 \\
   \texttt{MagFace} & 0.2034 & 0.1905 & 0.0989 & 0.1540 & 0.04054 & 0.1353 & 2.056 \\
   \texttt{CIFP} &  \textbf{0.1683} &  \textbf{0.1730} &  \textbf{0.0970} &  \textbf{0.1293} & 0.03097 & 0.1175 & 1.78 \\
   \texttt{MixFairFace} & 0.4661 & 0.2869 & 0.2928 & 0.3155 & 0.07349 & 0.1032 & 1.627 \\
   %\texttt{Ours-Ratial} & \textbf{0.09066} & \textbf{0.9683} & 0.1888 & 0.1844 & 0.1163 & 0.1495 & 0.02934 & 0.09877 & 1.623\\
   \texttt{Proposed} & 0.1810 & 0.1871 & 0.1163 & 0.1625 & \textbf{0.02775} & \textbf{0.08922} & \textbf{1.609} \\
   \bottomrule
  \end{tabular}
\end{table*}

\begin{table*}[t]
 % \caption{RFWを用いたラベルあり学習に関する公平性評価，およびLFWを用いたラベルなし学習（未知属性）に関する公平性評価結果：LFWはサンプル数が100個以上の26属性を評価対象とし，各属性でユーザを分割した際の識別性能の偏りを評価した}
 \caption{The performance and fairness evaluation results trained on BUPT-Balancedface dataset and evaluated on LFW dataset: For the LFW dataset, 26 attributes with more than 100 samples each were selected as the subjects for evaluation. STD, Gini, SER were assessed when users were divided according to these attributes, respectively.}
 \label{table:fairness}
 \centering
  \scalebox{1.0}{
      \begin{tabular}{cccccc}
      \toprule
       &EER($\downarrow$) & AUC($\uparrow$) & STD($\downarrow$) & Gini($\downarrow$) & SER($\downarrow$)\\
       \midrule
       \texttt{ArcFace} & 0.09300 & 0.9665 & 0.01170 & 0.08292 & 2.766 \\
       \texttt{MagFace} & 0.09867 & 0.9590  & 0.01127 & 0.08279 & 2.766 \\
       \texttt{CIFP} &\textbf{0.09100} & 0.9614 & 0.01157 & 0.08845 & 3.038\\
       % \texttt{Ours-Racial} & 0.01193 & 0.09035 & 2.640 \\
       \texttt{Proposed} & \textbf{0.09100} & \textbf{0.9681} & \textbf{0.01019} & \textbf{0.07398} & \textbf{2.525} \\
       % \texttt{MixFairFace} & 0.16900 & 0.9033 & 0.01667 & \textbf{0.06055} & \textbf{1.850} \\
       \bottomrule
      \end{tabular}
  }
\end{table*}
%
% Mean:  0.07320164050765356
% Max:  0.029413801865015646
% Min:  -0.039423844825637405
% Gini:  8.844852997789365
% STD:  0.011571461449321302
% SER:  3.0379555652088728

\subsection{Results}
% 我々の実験では，提案手法を以下の手法と比較した，1)\texttt{ArcFace}\cite{}：最も基本的な顔表現の学習手法，2) \texttt{MagFace}\cite{}：データ品質を考慮した顔表現の学習手法, 3) \texttt{CIFP}\cite{}：異なる人種間の偽陽性率の不一致を最小化させる手法, 4) \texttt{MixFairFace}：個人間における特徴距離を均一化させる手法, 
% %5) \texttt{Ours-Racial}：人種間における精度を均一化させる手法，
% 5) \texttt{Proposed}：個人間における確信度を均一化させる手法．
% LFWおよびRFWを用いた既知属性に関する性能評価および公平性評価結果をTable\ref{table:performance}に示す．また，LFWを用いた未知属性に関する公平性評価結果をTable\ref{table:fairness}に示す．
In our experiments, we compared the proposed method with the following approaches: 
(1) \texttt{ArcFace} \cite{arcface}: A basic method for learning facial representations,
(2) \texttt{MagFace} \cite{magface}: A method for learning facial representations considering sample quality,
(3) \texttt{CIFP} \cite{labeling_dataset}: A method to minimize the disparity in false positive rates across different races,
(4) \texttt{MixFairFace} \cite{mixfairface}: A method to equalize feature distances among individuals, and 
(5) \texttt{Proposed}: A method to equalize confidence level among individuals.

%図\ref{fig:roc_det}より，ROC曲線，DET曲線ともに\texttt{MixFairFace}をのぞいて各モデル間の性能評価差は小さいことがわかる．

% ROC/DETは抜く 2024/04/30 Ohki
% \begin{figure*}[b]
%   \begin{minipage}[b]{0.49\linewidth}
%     \centering
%     \includegraphics[keepaspectratio, scale=0.41]{fig/roc.pdf}
%     \subcaption{ROC curve}
%   \end{minipage}
%   \begin{minipage}[b]{0.49\linewidth}
%     \centering
%     \includegraphics[keepaspectratio, scale=0.41]{fig/det.pdf}
%     \subcaption{DET curve}
%   \end{minipage}
%   \caption{The performance evaluation of each model on LFW.}
%   \label{fig:roc_det}
% \end{figure*}
\subsubsection{Fairness-Accuracy Trade-off}
\label{subsec:eval_performance}
% % なお，\texttt{Ours-Racial}は人種間の公平性を考慮した提案手法であり，\ref{subsec:class_favoritism_level}節において，人種ごとにクラス優遇度を算出する．\texttt{Ours-Identity}は個人間の公平性を考慮した提案手法であり，\ref{subsec:instance_favoritism_level}節において，個人ごとにクラス優遇度を算出する．
% % LFWを用いたEER，AUCの結果を表\ref{table:performance}に，ROC曲線，DET曲線を図\ref{fig:roc_det}に示す．
% BUPT-Balancedfaceで学習を行い，LFWおよびRFWで性能評価を行った．LFWのEER,AUCおよびRFWの各属性に対するEERを表\ref{table:performance}に示す．Africanにおける各人種ごとのEERをそれぞれEER-Af(African)，EER-As(Asian)，EER-Ca(Caucasian)，EER-In(Indian)と表記する．表\ref{table:performance}より，LFWに対しては\texttt{Proposed}が最も性能が高い性能を示した．一方，RFWにおける各属性に関するEERはCIFPの性能が高いが，これはCIFPがあらかじめラベル付された人種情報を考慮した学習アルゴリズムを用いていることに起因すると考えられる．なお，MixFairFaceについては，著者らの実装および論文中のパラメータを用いて再現実験を実施したが，高い性能が得られなかった．
% Our performance evaluation results trained on BUPT-Balancedface dataset and evaluated on the LFW and RFW datasets are shown in Table \ref{table:performance}. EER for each race is denoted as EER-Af (African), EER-As (Asian), EER-Ca (Caucasian), and EER-In (Indian). As shown in Table \ref{table:performance}, the \texttt{Proposed} method achieved the highest performance on LFW. On the other hand, \texttt{CIFP} achieved the highest performance in EER across various attributes on RFW, which is believed to be due to \texttt{CIFP} utilizing a training algorithm that takes into account pre-labeled racial information. As for \texttt{MixFairFace}, despite conducting replication experiments using the implementation and parameters published by the authors of original paper\footnote{\url{https://github.com/fuenwang/MixFairFace}}, we could not achieve high performance.

Our study investigated whether the proposed method could improve fairness with respect to sensitive attributes (e.g., race or gender) while maintaining accuracy. Our performance and fairness evaluation results trained on BUPT-Balancedface dataset and evaluated on the RFW datasets are shown in Table \ref{table:performance}. EER for each race is denoted as EER-Af (African), EER-As (Asian), EER-Ca (Caucasian), and EER-In (Indian). As shown in Table \ref{table:performance}, \texttt{CIFP} achieved the highest performance in EER across all racials on RFW, which is believed to be due to \texttt{CIFP} utilizing a training algorithm that takes into account pre-labeled racial information. In contrast, the \texttt{Proposed} method exhibits a lower (the best) STD/Gini/SER values compared to other methods. The differences of EERs between  \texttt{Proposed} and \texttt{ArcFace} are small, indicating that  \texttt{Proposed} improves fairness while maintaining authentication accuracy. As for \texttt{MixFairFace}, despite conducting replication experiments using the implementation and parameters published by the authors of original paper\footnote{\url{https://github.com/fuenwang/MixFairFace}}, we could not achieve high performance.

\subsubsection{Label-Independent Fairness Improvement}
\label{subsec:eval_fairness}
\begin{figure*}[htbp]
\centering

% 上段の3つの図
\begin{subfigure}{.45\textwidth}
  \centering
  \includegraphics[width=1.1\linewidth]{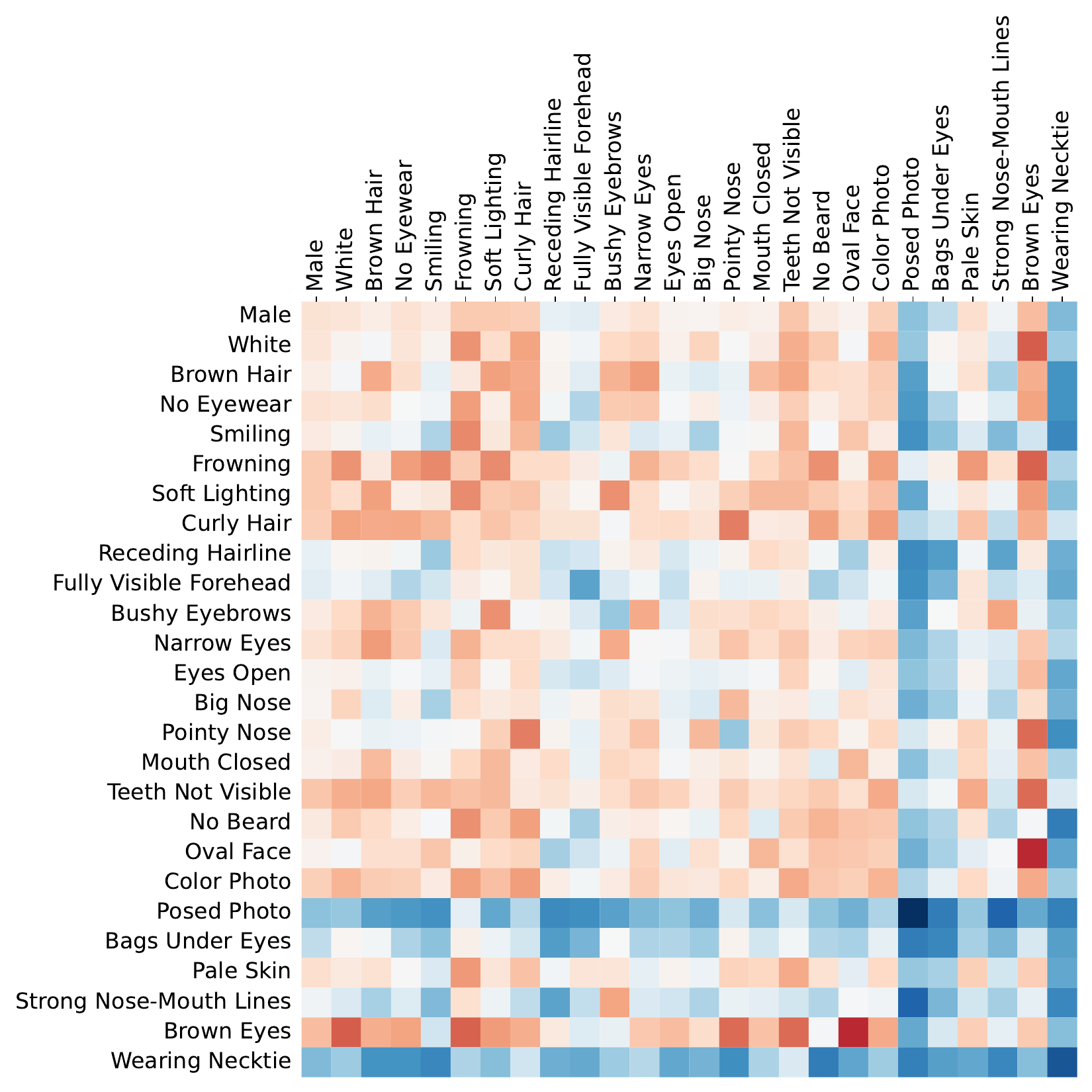}
  \caption{\texttt{ArcFace} (STD=0.01170)}
\end{subfigure}%
\hspace{.05\textwidth}
\begin{subfigure}{.45\textwidth}
  \centering
  \includegraphics[width=\linewidth]{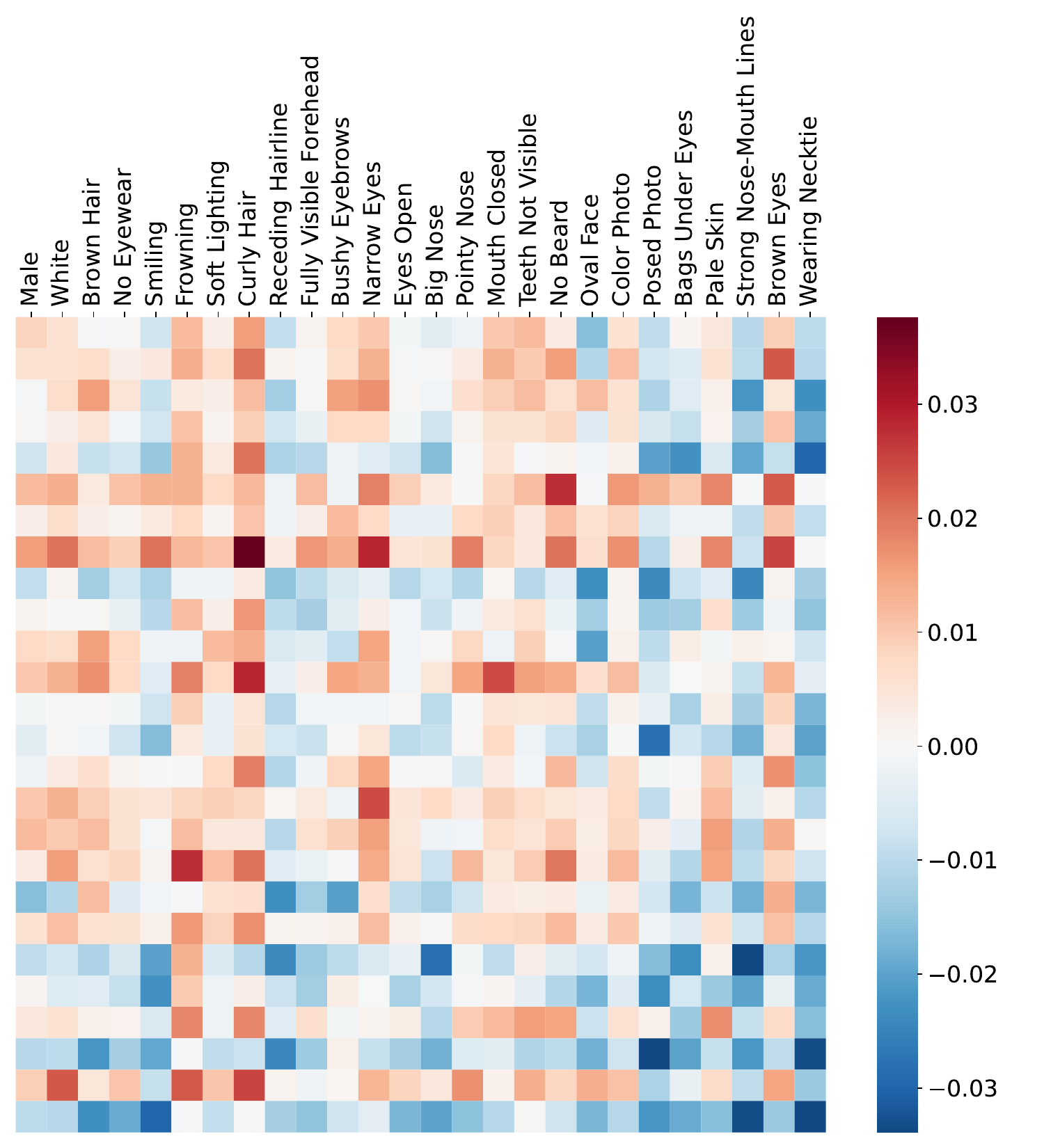}
  \caption{\texttt{MagFace} (STD=0.01127)}
\end{subfigure}%
\vspace{1em} % 上下の図の間隔を調整

% 下段の2つの図
\begin{subfigure}{.45\textwidth}
  \centering
  \includegraphics[width=1.1\linewidth]{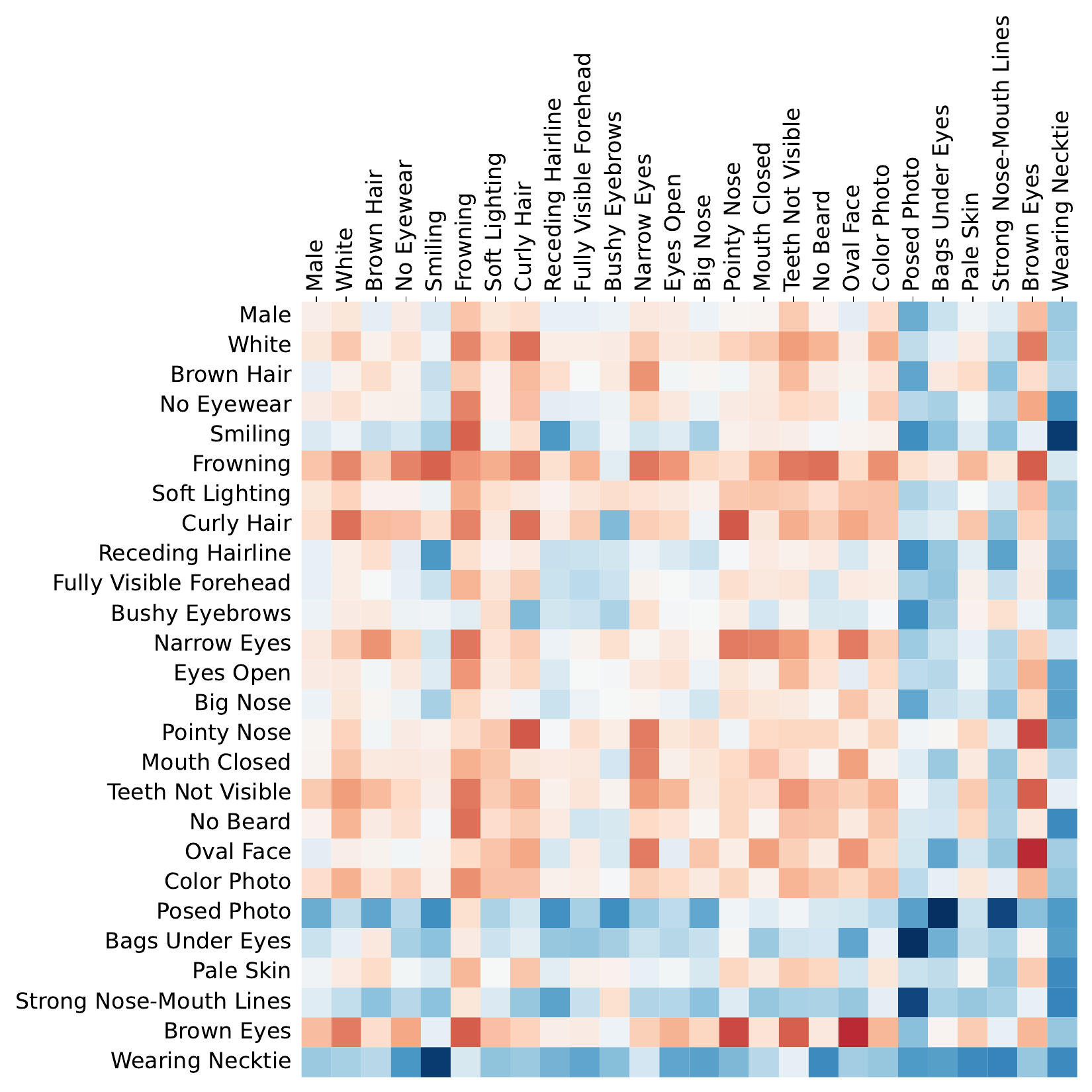}
  \caption{\texttt{CIFP} (STD=0.01157)}
\end{subfigure}
% \begin{subfigure}{.32\textwidth}
%   \centering
%   \includegraphics[width=\linewidth]{fig/heatmap_mixfairface_cut.pdf}
%   \caption{\texttt{MixFairFace} (STD=0.01667)}
% \end{subfigure}%
\hspace{.05\textwidth}
\begin{subfigure}{.45\textwidth}
  \centering
  \includegraphics[width=\linewidth]{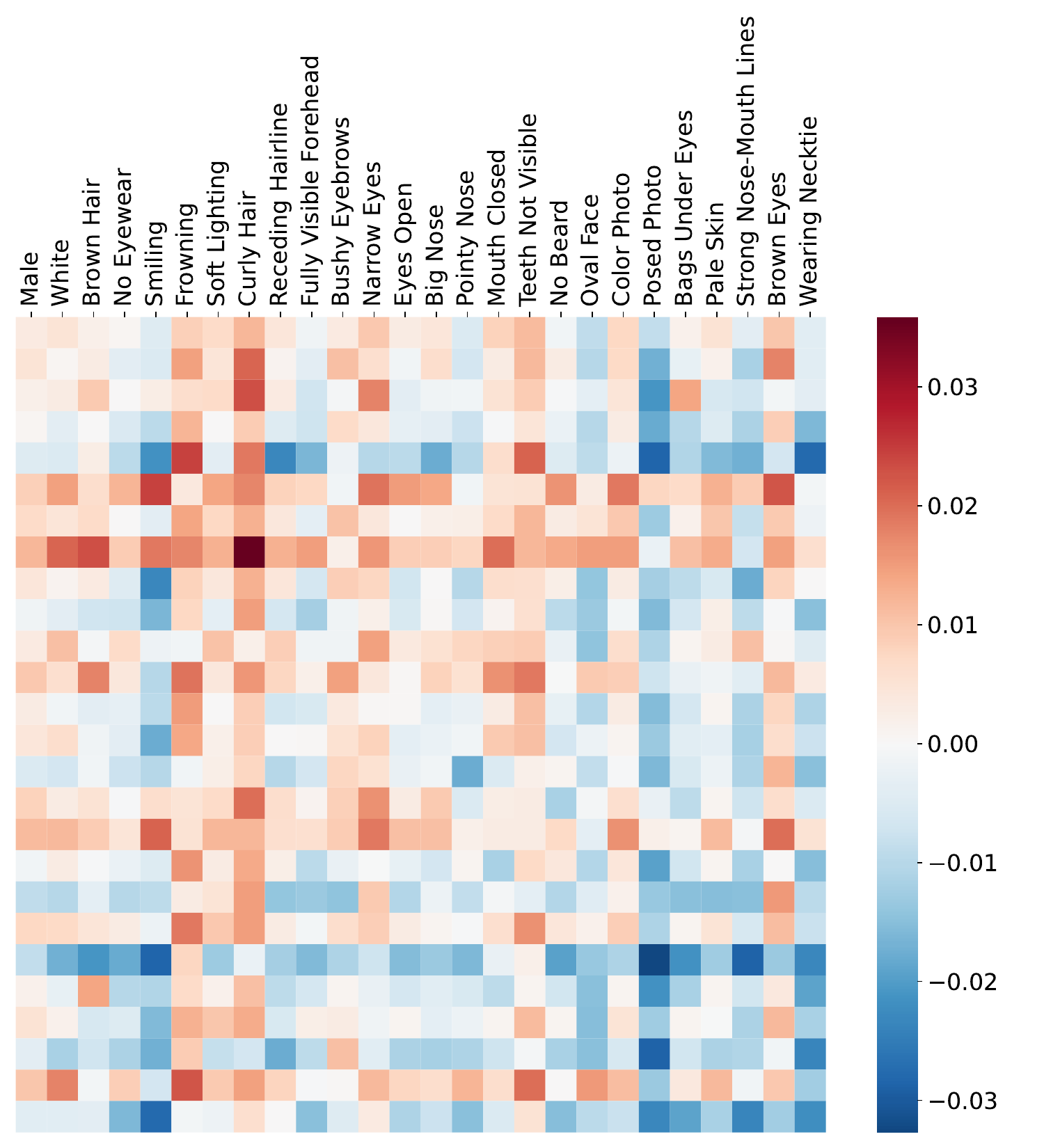}
  \caption{\texttt{Proposed} (STD=0.01019)}
\end{subfigure}

\caption{The fairness heatmap of each model across 26 attributes on LFW: Each cell indicates the deviation of EER, with blue indicating lower EER than the average and red indicating higher EER than the average. For reference, we include the values of the standard deviation (STD) from Table \ref{table:fairness} in parentheses.
}
\label{fig:confusion}
\end{figure*}

% BUPT-Balancedfaceデータセットで学習を行い，RFWを用いた人種間公平性のおよびLFWデータセットを用いて提案手法の人種間公平性について評価を行った．RFWおよびLFWにおけるSTD,Gini,SERの3尺度による公平性評価結果を表\ref{table:fairness}に示す．またLFWの26属性ドメインにおける公平性比較を図\ref{fig:confusion}にヒートマップとして示す．
% なお，LFWデータセットを用いた評価では，モデルの学習に用いていない未知の属性に対する公平性を評価するために，サンプルが100個以上存在する26属性を評価対象として．各属性でユーザを分割した際の識別性能の偏りを評価した．LFWの計74属性は連続値でラベル付けされており，大きい値であるほどその属性を含んでいる\cite{lfwa}．そこで連続値を[-1,1]にMinMaxスケーリングし，0.5以上の値であれば該当する画像がその属性を含んでいるものとした．
Next, our study explored whether our proposed method could improve fairness independent of annotated labels. Fairness evaluation results trained on BUPT-Balancedface dataset and evaluated on the LFW dataset are shown in Table \ref{table:fairness}. Additionally, a comparison of fairness across the 26 attribute domains in LFW is shown as a heatmap in Figure \ref{fig:confusion}. For the evaluation with the LFW dataset, 26 attributes with more than 100 samples each were selected for analysis. STD, Gini, SER were assessed when users were divided according to these attributes, respectively. The 74 attributes of LFW are labeled with continuous values, where higher values indicate a stronger presence of the attribute \cite{lfwa}. Hence, continuous values were MinMax scaled to the range $[-1,1]$, and values above 0.5 were considered to indicate the presence of the attribute in the images. \texttt{MixFairFace}, which did not perform well in section \ref{subsec:eval_performance}, was excluded here.

% 表\ref{table:fairness}より，MixFairFaceはLFWにおいて高いGini係数およびSERを示している．しかし，STDが大きいことからもわかるように，これはMixFairFaceの識別精度そのものが低いことに起因すると考えられる．また，CIFPは特にLFWにおいて他の手法と比較して公平性評価指標がいずれも悪い結果となっている．これはCIFPが人種のみを考慮した学習を行っているため，未考慮の属性に対する公平性が悪化したことを示唆している．これらに対して\texttt{Proposed}はSTDが小さく，またGini/SERも他手法と比較して良好な値を示している．これらの結果は，個人を単位とした公平性学習を行うことで，未知の属性に対しても高い精度と公平性のトレードオフを達成する学習が可能であることを示唆している．
As shown in Table \ref{table:fairness}, \texttt{CIFP} has shown poorer performance on all fairness metrics compared to other methods, suggesting that its fairness for attributes not considered in training has deteriorated because it focuses only on racial attributes. In contrast, the \texttt{Proposed} method performs best on all performance and fairness metrics. These results suggest that the label-free fairness training at an individual level proposed by this method can achieve a high trade-off between accuracy and fairness even for unknown attributes. 

% \subsection{Other Attributes Fairness Experiment}
% 図\ref{fig:confusion}，表\ref{table:fairness}よりLFWの学習時に未知である26属性の公平性は提案手法モデルがすべての指標で最も優れていることがわかった．

% 特に，人種考慮モデルはArcFaceと比較して，STDとGiniが悪い結果となった．これは，属性の中で人種ドメインのみを考慮しながら学習を行った結果，未知の属性ドメインの公平性が悪化したことを示唆している．

% 一方で，提案手法モデルはArcFaceよりも公平性指標において良い結果を示しているため，クラス単位でマージンを動的に変化させることが結果的に様々な属性ドメインにおいて公平性を向上させる要因となったと考えられる．

% \begin{figure*}[t]
%   \begin{minipage}[b]{0.35\linewidth}
%     \centering
%     \includegraphics[keepaspectratio, scale=0.49]{fig/heatmap_arc.pdf}
%     \subcaption{ArcFace Model}
%   \end{minipage}
%   \begin{minipage}[b]{0.30\linewidth}
%     \centering
%     \includegraphics[keepaspectratio, scale=0.49]{fig/heatmap_attr.pdf}
%     \subcaption{Racial Model}
%   \end{minipage}
%   \begin{minipage}[b]{0.33\linewidth}
%     \centering
%     \includegraphics[keepaspectratio, scale=0.49]{fig/heatmap_class.pdf}
%     \subcaption{Proposed Model (\texttt{Proposed})}
%   \end{minipage}
%   % ArcFace, MagFace, CIFP, MixFairFace, Proposed
%   \setlength{\captionmargin}{20pt}
%   \caption{The fairness heatmap of each model across 26 attributes on LFW.\\Each cell indicates the deviation of EER, with blue indicating lower EER than the average and red indicating higher EER than the average.}
%   \label{fig:confusion}
% \end{figure*}

\section{Discussions}
\noindent\textbf{Computational cost.}
%\ref{fig:overview_cfl}で示すClass Favoritism Levelの計算は，学習時に逐次的に$\overline{P_j}$を計算することで必要なメモリ容量を少量に抑えることができる．それでも，学習データ数に比例した計算量の増加が見込まれるので，この点については考慮する必要がある．
The calculation of the Class Favoritism Level, as shown in Figure \ref{fig:overview_cfl}, can keep the required memory capacity low by sequentially calculating $\overline{P_j}$ during training. However, since the computation increases in proportion to the number of training data, this aspect needs to be considered.

\noindent\textbf{Selection of Hyperparameters.} 
%本手法では公平性と精度のトレードオフを調整するパラメータとして$gamma$および$h$が存在する．本論文においては$h$を0から1，$\gamma$を1から20の範囲でグリッドサーチした結果最も性能が高い$h=1, \gamma=10$の組み合わせを用いている．$h$はその値が大きいほど優遇属性の学習を可能な限り抑えつつ不遇属性の学習を進める．また，$\gamma$はその強度の調整パラメータである．データセットにおける潜在的な属性偏りが大きいと予想される場合には，$h$や$\gamma$をより大きな値とすることで高い公平性改善効果が得られると考えられる．一方これらの最適な決定法については本論文では提案しておらず，今後の検討課題である．
In this method, the parameters $\gamma$ and $h$ are employed to balance the trade-off between fairness and accuracy. A grid search was performed for $h$ and $\gamma$, revealing that the best performance was achieved with $h = 1$ and $\gamma = 10$.
Higher values of $h$ suppress the learning of favored attributes while encouraging the learning of neglected attributes. $\gamma$ determines the intensity of this effect.
If significant latent attribute biases are expected in the dataset, it is suggested that using larger values for $h$ or $\gamma$ could lead to greater fairness improvements. However, it should be noted that excessively large values may cause instability in the learning process. This paper does not propose an optimal method for determining these parameters, leaving it as a topic for future research.

\section{Conclusion}
% 本研究では，顔認証における認証バイアス軽減を目的として新たなフレームワーク，LabellessFaceを提案した．本手法は人工統計学的な事前のラベル付けを必要としないラベルフリーの学習アルゴリズムの実現を目的とした．この目的を達成するために，我々は，クラス優遇度に基づくFair class margin penaltyの概念を提案し，最小のクラス単位である個人単位のクラス優遇度を用いることで，公平性考慮のための人工統計学的ラベル付けの必要を必要としない学習を提案した．一般的な顔ベンチマークを用いた広範な実験により他のベースラインと比較した本提案手法の有効性，特に，広範な属性における公平性を学習時に考慮することなしに実現可能である点を示した．今後はより最適なマージン係数の決定方法や，ハイパーパラメータの最適化など，様々な方面での研究の拡張が期待できる．
In this paper, we proposed a new framework, LabellessFace, aimed at reducing authentication bias in facial recognition. 
This method was designed to realize a label-free training method that does not require pre-labeling for demographic groups. 
To achieve this goal, we introduced the concept of a fair class margin penalty based on class favoritism levels, utilizing individual class units to avoid the need for demographic group labeling for fairness considerations. 
Extensive experiments using common facial benchmarks demonstrated the effectiveness of our proposed method compared to other baselines, particularly in achieving fairness across a broad range of attributes without the need for consideration during training. Future work can explore further research extensions in various aspects, such as determining more optimal margin coefficients and optimizing hyperparameters.

% 本論文では、顔認識における偏りを緩和し、公平性を向上させるために、ソフトマックス損失関数に新しいペナルティ項を開発する。我々は、人口統計FPRの極端なケースとしてインスタンスFPRの概念を提案し、インスタンスFPRの一貫性をsoftmaxベースの損失のペナルティ項目として変換する。一般的な顔ベンチマークを用いた広範な実験により、SOTA競合手法と比較した本手法の有効性を実証する。本研究の主旨を踏まえると、将来的には、より良い重み関数F (-)の設計や、偽陽性事例として誤って最適化される可能性のあるノイズサンプルの影響の調査など、様々な側面で研究を拡張することができる。

% クラスタリングされた属性ごとに適切な学習を行う公平性向上のための従来のアプローチとは異なり，個々のクラス（顔認証においては個人）の認証性能を均等にするように学習を行うことで，

% 評価実験の結果，ArcFaceの$EER=0.093$に対して提案手法モデルは$EER=0.091$を示し，認証精度が向上した．さらに，人種間や様々な属性間の公平性は，三つのモデルの中で提案手法モデルが最も優れていた．

% 人種間の公平性評価において，ArcFaceの$Gini=10.31,SER=1.725$に対して人種考慮モデルは$Gini=9.877,SER=0.1623$を示し，ふたつの指標ともに公平性が向上した．

% この結果は，学習データの人種情報が不正確であり，人種情報を自動的にラベリングすることの困難性を示唆している．

% さらに，学習時に考慮していない未知属性間の公平性評価において，ArcFaceの$Gini=8.279,SER=2.766$に対して人種考慮モデルは$Gini=9.035,SER=2.64$と，公平性が向上したとは言えない結果となった．

% これは，学習時に考慮していない属性に対しては公平性が向上しないことを示唆している．

% 一方で，提案手法モデルは未知属性間の公平性評価において，$Gini=8.922,SER=1.609$と，三つのモデルの中で最も公平なモデルであることが示された．

% 本実験で用いたLFW,RFWによる性能評価は比較的簡単なタスクであり，近年の高品質なアーキテクチャや学習データセットを利用するとAccuracyが0.999\cite{paperwithcode}と非常に高い値となってしまい，提案手法の有効性が評価することが難しい．
% そこで，提案手法の有効性を明確にするため，モデルは比較的軽量なアーキテクチャであるResNet34を用いたり，学習データセットも比較的小さめなBUPT-Balancedfaceを使用している．

% このような実験設定から，本結果を用いて近年のSOTAモデルと比較評価することはできない．

% そのため，今後は運用場面に相当するモデルや学習データ，学習設定において提案手法を用いることや，他の様々な公平性改善手法との比較を行い，先行研究との位置関係を明らかにしたい．

\section*{Acknowledgement}
This work was supported in part by JSPS KAKENHI JP 23K28085, and JST Moonshot R\&D Grant Number JPMJMS2215.

{\small
\bibliographystyle{ieee}
\bibliography{references}
}

\end{document}